\documentclass[conference]{IEEEtran}
\IEEEoverridecommandlockouts
\usepackage{subfig}
\usepackage{dsfont}
\usepackage{enumitem} 
\usepackage{multirow}
\usepackage{amsthm}
\usepackage{tikz}
\usepackage{pgfplots}
\usepgfplotslibrary{groupplots}

\usepackage{cite}
\usepackage{amsmath,amssymb,amsfonts}
\usepackage{algorithmic}
\usepackage{graphicx}
\usepackage{textcomp}

\usepackage{xcolor}
\pgfplotsset{compat=1.18}

\def\BibTeX{{\rm B\kern-.05em{\sc i\kern-.025em b}\kern-.08em
    T\kern-.1667em\lower.7ex\hbox{E}\kern-.125emX}}
\begin{document}

\title{PNCS:Power-Norm Cosine Similarity for Diverse Client Selection in Federated Learning\\

}


\author{
\IEEEauthorblockN{Liangyan Li\IEEEauthorrefmark{1}, Yangyi Liu\IEEEauthorrefmark{1}, Yimo Ning\IEEEauthorrefmark{1}, Stefano Rini\IEEEauthorrefmark{2}, Jun Chen\IEEEauthorrefmark{1} }
\IEEEauthorblockA{\IEEEauthorrefmark{1}Department of Electrical and Computer Engineering, McMaster University, Hamilton, Canada \\
Emails: \{lil61, ningy7, chenjun\}@mcmaster.ca, yangyiliu21@gmail.com}
\IEEEauthorblockA{\IEEEauthorrefmark{2}Department of Electrical and Computer Engineering, National Yang-Ming Chiao-Tung University, Taiwan \\
Email: stefano@nycu.edu.tw}
}


\maketitle

\begin{abstract}
Federated Learning (FL) has emerged as a powerful paradigm for leveraging diverse datasets from multiple sources while preserving data privacy by avoiding centralized storage. However, many existing approaches fail to account for the intricate gradient correlations between remote clients, a limitation that becomes especially problematic in data heterogeneity scenarios. In this work, we propose a novel FL framework utilizing Power-Norm Cosine Similarity (PNCS) to improve client selection for model aggregation. By capturing higher-order gradient moments, PNCS addresses non-IID data challenges, enhancing convergence speed and accuracy. Additionally, we introduce a simple algorithm ensuring diverse client selection through a selection history queue. Experiments with a VGG16 model across varied data partitions demonstrate consistent improvements over state-of-the-art methods.

\end{abstract}


\section{Introduction}
\label{sec:introduction}

Foundation models, such as large language models, benefit from the diverse information available in decentralized client data. To safeguard user privacy, Federated Learning (FL) has emerged as a collaborative paradigm that trains models without sharing raw data~\cite{mcmahan2017federated, li2020convergencefedavgnoniiddata}. In this setup, a central \emph{parameter server} (PS) coordinates global model training by aggregating updates from distributed clients, each maintaining the privacy of its local dataset. 
%
Federated Learning (FL) faces challenges of local data heterogeneity, which hinders model convergence, and communication inefficiency due to frequent exchanges of updates between clients and the central server~\cite{konevcny2016federated, li2020convergencefedavgnoniiddata}.
%

This paper addresses the intersection of these two issues. In particular, we seek to answer the following question:

\emph{``What is the best strategy for selecting a subset of clients for training under local data heterogeneity in terms of model convergence speed and communication cost?''}

Our core insight differs from existing client-selection literature in two key respects. First, we highlight that \emph{diversity} among the selected clients, rather than strict gradient alignment, is often the more effective criterion, as divergent gradients can provide complementary updates for the global model. Second, we propose Power-Norm Cosine Similarity (PNCS) as a particularly robust selection metric under various data heterogeneity levels. 

Our contributions can be outlined as follows:
\begin{itemize}[leftmargin=*]
    \item \textbf{Novel Problem Formulation for Client Selection:} 
    We formulate the client selection problem in FL as a logistic regression model in which the input features capture the pairwise gradient diversity between any two clients, and the output is the probability of how likely the pair may be the best choice for model update. This notion naturally extend to the multi-client case by averaging the pairwise selection performance.
    \item \textbf{Feature Exploration for Gradient Alignment:} 
    We thoroughly evaluate a wide range of features for two-client selection in a single-layer FL training scenario, including both statistical and geometrical criteria. Our analysis, conducted across varying number levels of data heterogeneity and number of selected clients, reveals that the cosine similarity induced by the $L_K$ norm (with $K \in \{1,2,4\}$) performs robustly across all scenarios.

    \item \textbf{Identification of PNCS as the Best Single Feature:} 
    Among the different $L_{K}$-based cosine similarity measures, we identify $L_4$ as the most effective single feature for client selection, as it better captures gradient alignment under diverse training conditions. We refer to our client selection methods in FL as PNCS.
    \item \textbf{Empirical Validation and Performance Gains:} 
    We conduct extensive numerical evaluations demonstrating that leveraging PNCS for client selection significantly improves convergence speed and final model accuracy compared to existing selection strategies. 
\end{itemize}

\section{Related Work}
\label{sec:related_work}


Client selection in FL is motivated by various local dataset dynamics such as data imbalance, heterogeneity, and computational constraints, as established in foundational works~\cite{mcmahan2017communication,ma2022state,li2024comprehensive}. 
Research has shown that strategic client selection can significantly improve accuracy~\cite{lai2021oort}, ensure fairness~\cite{sultana2022eiffel}, enhance robustness~\cite{balakrishnan2022diverse}, and accelerate convergence~\cite{nguyen2020fast,sultana2022eiffel} 
Lyapunov optimization frameworks~\cite{9283151,10219820,battiloro2022lyapunov,zhou2023joint} dynamically manage system stability across these dimensions, while greedy algorithms~\cite{mehta2023greedy,9317862,liu2023mflces} and Hungarian matching methods~\cite{mohammed2020budgeted,chen2019pre} offer more computationally efficient solutions. Reinforcement learning approaches~\cite{9372789,albelaihi2023deep,arulkumaran2017deep} also allow adaptive selection policies to be learned from environmental interactions.

On the other hand, single-factor optimizations focus on isolated aspects like computational capacity~\cite{9991371,zhu2022online} or client reputation~\cite{wang2020novel,batool2023block,wang2021collective,tan2022reputation}, potentially overlooking the comprehensive trade-offs required for optimal global performance. 
The AFL method~\cite{afl} assigns selection probabilities via differential privacy applied to local losses, and POWD~\cite{powd} prioritizes clients with high local losses. While these loss-based methods can speed up convergence, they may degrade overall performance in non-IID settings by ignoring correlations among clients.

Other methods select clients by inspecting their gradients~\cite{marnissi2024client,sivasubramanian2024gradient,xu2023federated}. In particular, \cite{marnissi2024client} selects clients whose gradients have the largest norms, and \cite{sivasubramanian2024gradient} leverages Shapley values to identify clients most representative of the global dataset, \cite{xu2023federated} jointly optimizes client selection and gradient compression. However, these gradient-based methods typically treat each client's contribution independently, risking biased updates when chosen clients fail to represent the global distribution.  FedCor~\cite{fedcor} models correlations between clients’ losses via a Gaussian Process, allowing it to iteratively select clients based on predicted performance gains. 

\section{Preliminaries}

\subsection{Notations}
\label{sec:notations}

We use lowercase boldface letters (e.g., $\mathbf{z}$) to denote vectors and calligraphic uppercase letters (e.g., $\mathcal{A}$) to denote sets. For any set $\mathcal{A}$, let $|\mathcal{A}|$ represent its cardinality. We adopt the shorthand notation $[m:n] \triangleq \{m, \ldots, n\}$ and $[n] \triangleq \{1, \ldots, n\}$. Throughout the paper:
 $k \in [K]$: Represents a client index, with $k'$ used for two client indexes.
 $t \in [T]$: Denotes an iteration index.
$s \in \mathcal{S} \subseteq [K]$: Represents a selected client index, where $|\mathcal{S}| = S$.
 $j \in [J]$: Denotes a shard (or partition) index.
 $r \in [R]$: Indicates the random seed.

\subsection{Federated Learning}
\label{sec:preliminary}

Consider the FL setting with $K$ clients, each possessing a local dataset
${\mathcal{D}}_k \in \mathcal{D}$, 
for $k \in [K]$, wishing to minimize the \emph{loss function} $\mathcal{L}$ as evaluated across all the clients and over the model weights $\mathbf{w} \in \mathbb{R}^m$, where $m$ denotes the dimensionality of the model parameters. 

This minimization is coordinated by the PS as follows: in round $t \in [T]$, the clients transmit local gradients to the PS; the PS generates a model update and sends the updated model back to the clients. 
These steps are repeated $T$ times, and the model obtained at time $T$ is declared as the converged model. 

Mathematically, the loss function $\mathcal{L}$ is defined as
\begin{equation}
    \mathcal{L}(\mathbf{w}) = \frac{1}{|\mathcal{D}|} \sum_{k \in [K]}  {\mathcal{L}}_k({\mathcal{D}}_k, \mathbf{w}),
\label{eq:loss}
\end{equation}
where ${\mathcal{L}}_k({\mathcal{D}}_k, \mathbf{w})$ is the local loss function quantifying the prediction error of the $k$-th client's model. A common approach for numerically finding the optimal value of $\mathbf{w}$ is through the iterative application of (synchronous) stochastic gradient descent (SGD). 

We define the local gradients calculated at communication round $t$ as
\begin{equation}
    \mathbb{E}[\mathrm{g}_{kt}] = \mathbb{E} [ \nabla {\mathcal{L}}_k({\mathcal{D}}_k, \mathbf{w}_t)],
\label{eq:gradient}
\end{equation}
where $\nabla {\mathcal{L}}_k({\mathcal{D}}_k, \mathbf{w}_t)$ denotes the local gradients of the model evaluated at the local dataset of the $k$-th client by minimizing the local loss function.
Note that the expectation in \eqref{eq:gradient} is taken over the randomness in evaluating the gradients, e.g., mini-batch effects.
The PS aggregates all the local gradients and forms the new global weights:
$    \mathbf{w}_{t+1} =  \mathbf{w}_{t} - \eta_{t} {\mathrm{\mathbf{g}}_t},$ for $ t \in [T]$,  where  
 ${\mathrm{\mathbf{g}}_t} = \frac{1}{K} \sum_{k \in [K]} \mathrm{\mathbf{g}}_{kt}$, $\mathbf{w}_0$ is a random initialization.

\subsection{Client Selection}
\label{sec:client_selection}

Communication overhead is a major bottleneck in FL, as each training round typically involves transmitting updates between multiple clients and a central PS. To address this, a common strategy is to select only a subset of clients in each round, reducing communication while still gathering sufficiently diverse updates. Formally, let $\mathcal{S}_t \subseteq [K]$ denote the set of active clients at round $t \in [T]$. This selection process aims to balance two objectives: keeping $|\mathcal{S}_t|$ small to mitigate communication costs, and ensuring that the selected gradients are sufficiently diverse to enhance the global model’s generalization. Client selection can follow two main paradigms. In \emph{decentralized} approaches, each client independently decides whether to participate based on local metrics. In contrast, \emph{centralized} approaches place this responsibility on the PS, which can use partial gradient statistics for client selection. 

In this work, we focus on centralized selection under heterogeneous data distributions, leveraging gradient-based measures to identify and activate clients whose updates offer the greatest potential for improving model convergence.

\subsection{Data Heterogeneity}
\label{sec:data_heterogeneity}

In many practical scenarios, the local dataset at each client is intrinsically heterogeneous. One way to capture this heterogeneity is by assuming that each client's data are sampled from a mixture of several prior distributions. 

Specifically, let $\mathbf{m}_{ik}$ denote the $i$-th data point (features and labels) at client $k$. We write
\begin{equation}
\mathbf{m}_{ik} \,\sim\, P_{\mathbf{m}}^k \;=\; \sum_{j \in [J]} \lambda_{jk}\,P_{\mathbf{m}}^{(j)},
\label{eq:mixture}
\end{equation}
where $\{P_{\mathbf{m}}^{(j)}\}_{j \in [J]}$ is a kernel of $J$ distributions, and the mixing coefficients $\lambda_{k} = \{\lambda_{jk}\}_{j \in [J]}$ specify the proportion of data at client $k$ coming from each kernel distribution.

\section{Problem Formulation}
\label{sec:Problem Formulation}

In many practical FL scenarios, it is advantageous for the PS to decide which clients will actively participate in a training round, based on partial information about each client’s local gradients. 
We refer to this approach as the \emph{Gradient Summaries for Centralized Client Selection} (GSCCS) setting.
The selection process proceeds in two phases:
\begin{enumerate}[leftmargin=*]
    \item \textbf{Summary Transmission:} Each client $k \in [K]$ sends a \emph{gradient summary}—a compressed or otherwise optimized representation of its local gradient $\mathbf{g}_{kt}$—to the PS.
    \item \textbf{Centralized Client Selection:} Upon receiving these summaries, the PS examines their contents and decides on the set $\mathcal{S}_t \subseteq [K]$ of active clients for round $t$, with $|\mathcal{S}_t| = S$.
\end{enumerate}

By introducing this extra summary-transmission phase, the PS can make a more informed selection of active clients. Compared to one-shot aggregation schemes that skip client selection, GSCCS adds one extra transmission step prior to the main gradient exchange. However, these summaries typically have much lower dimension or precision than the full gradient vectors, allowing the PS to \emph{exclude} clients whose updates do not significantly benefit the global model.

Mathematically, let the \emph{gradient summary} of client $k$ at time $t$ be defined as 
\begin{equation}
\mathbf{s}_{kt} = \phi(\mathbf{g}_{kt}),
\label{eq:summary_def}
\end{equation}
where $\phi: \mathbb{R}^m \to \mathbb{R}^p$ outputs a lower-dimensional or compressed vector. Each client $k \in [K]$ transmits $\mathbf{s}_{kt}$ to the PS, which then uses these summaries to compute pairwise gradient affinities and ultimately select the active set $\mathcal{S}_t$ of cardinality $J$. Once $\mathcal{S}_t$ is determined, the gradient update at round $t$ is given by
\begin{equation}
\hat{\mathbf{g}}_t = \frac{1}{J} \sum_{s \in \mathcal{S}_t} \mathbf{g}_{st},
\label{eq:aggregation_2}
\end{equation}
and the corresponding sequence of model weights $\{\hat{\mathbf{w}}_t\}_{t \in [T]}$ follows from standard gradient-based updates. We denote the client selection policy as
\begin{equation}
\mathcal{S}_t = \pi\bigl(\{\mathbf{s}_{kt}\}_{k \in [K]}\bigr),
\end{equation}
where among the $\binom{K}{J}$ possible subsets of clients, the \emph{optimal} choice $\mathcal{S}_t^*$ minimizes the loss $\mathcal{L}(\cdot)$ at time $t$. Thus, the client selection problem under the GSCCS setting can be formulated as
\begin{equation}
\mathcal{P}: \quad \min_{\phi, \pi} \quad \frac{1}{T} \sum_{t \in [T]} \left[ \mathcal{L}\bigl(\hat{\mathbf{g}}_t^*\bigr) \;-\; \mathcal{L}(\hat{\mathbf{g}})\right],
\label{eq:prob_definition}
\end{equation}
where we seek to minimize the average additional loss incurred by selecting $J$ out of $K$ clients based on summaries of dimension $p$.
\subsection{Assumptions, Observations, and Comments}
\label{sec:assumptions_observations}

\noindent
\textbf{Sketch dimension.}
We do not delve deeply into how the sketch size $r$ affects training performance, leaving this investigation for future work. Our current framework assumes $r$ is large enough that each sketch accurately captures the relevant gradient information needed for client selection.

\noindent
\textbf{Reliability of summaries.}
We assume the transmitted summaries are sufficiently accurate for the PS to compute meaningful gradient-affinity measures. In practice, one can optimize the sketching or sampling procedure (e.g., choosing an appropriate matrix $M$) to balance communication overhead against selection accuracy.

\noindent
\textbf{Randomness in the training process.}
When referring to ``optimal'' client selection, we do so in an \emph{average} sense—averaging over the stochasticity introduced by random shard allocations, parameter initialization, and mini-batch sampling. Analyzing the full dynamics of this random process is beyond our current scope.

\noindent
\textbf{Final-loss selection criterion.}
Selecting clients to directly minimize the \emph{final} loss after $T$ rounds can be impractical, as it requires predicting the model’s future evolution across multiple rounds. Instead, our approach focuses on a per-round selection strategy, striking a practical balance between simplicity and performance.

\section{Proposed Method: PNCS}
\label{sec:Proposed_Method}

In this section,
Section~\ref{sec:Logistic_Regression_Pairwise} introduces a logistic regression formulation for selecting \emph{exactly two} clients. Section~\ref{sec:Numerical_Findings} provides numerical insights into the most predictive gradient-based features, and Section~\ref{sec:cosl4-select} describes the full PNCS algorithm.

\subsection{Logistic Regression for Pairwise Client Selection}
\label{sec:Logistic_Regression_Pairwise}
We model selecting two clients from $K$ as a logistic regression problem, using input features to evaluate their interaction and labels based on normalized validation accuracies, creating a dataset of feature-label pairs across various settings.

Mathematically, we define the \emph{client selection} (CS) loss:
\begin{equation}
\label{eq:log_loss}
\mathcal{L}_{CS} 
= 
\min_{\mathbf{w}} 
\frac{1}{J} 
\sum_{\substack{k,k' \\ k \neq k'}} 
\mathsf{BCE}\!\Bigl(\sigma(\mathbf{w}^\top \mathbf{x}_{k,k'}), \,y_{k,k'}\Bigr),
\end{equation}
where \(\mathsf{BCE}\) is the binary cross-entropy loss, \(\sigma(\cdot)\) is the sigmoid function, \(\mathbf{x}_{k,k'}\) is a $d$-dimensional feature vector (e.g., pairwise gradient metrics) for clients $(k,k')$, and \(y_{k,k'} \in [0,1]\) reflects how well the pair \((k,k')\) performs if selected. Intuitively, \(\mathcal{L}_{CS}\) measures how accurately the logistic model (parameterized by \(\mathbf{w}\)) predicts the performance of any two-client combination under the specified training conditions.

\noindent
\textbf{Feature Vector Construction.}
In our experiments, we collect various gradient-based metrics into the feature vector \(\mathbf{x}_{k,k'}\) (see Table~\ref{tab:log reg all feat}). One key candidate is the \emph{cosine similarity} under the $L_p$ norm:
\begin{equation}
\cos_p(\mathbf{g}_k,\mathbf{g}_{k'}) 
\;=\; 
\frac{\langle \mathbf{g}_k, \mathbf{g}_{k'} \rangle_p}{\|\mathbf{g}_k\|_p \,\|\mathbf{g}_{k'}\|_p},
\end{equation}
with 
\begin{equation}
\langle \mathbf{u}, \mathbf{v} \rangle_p 
= 
\frac{\|\mathbf{u} + \mathbf{v}\|_p - \|\mathbf{u} - \mathbf{v}\|_p}{4}.
\end{equation}
This generalization encompasses the traditional $L_2$ cosine similarity as a special case and allows for higher-order norms (e.g., $L_4$) that can emphasize dominant gradient coordinates.

\noindent
\textbf{Label Definition.}
To generate the label \(y_{k,k'}\) in \eqref{eq:log_loss} for a particular pair of clients $k,k'$, we:
\begin{itemize}[leftmargin=*]
    \item Let \(y'_{k',k}(n,j,r)\) represent the vector of accuracies (or losses) for all possible $K(K-1)$ pairs under the three training conditions: (i) iteration \(n\), (ii) heterogeneity level \(j\), and (iii) random seed \(r\).
    \item Map these raw accuracies to \([0,1]\) using, for instance, a softmax or min-max scaling to obtain \(y_{k,k'}\). Other scaling functions can be considered, such as linear or entropy scaling.
\end{itemize}
The trained logistic model provides a weight vector $\mathbf{w}$ identifying the most predictive pairwise features for client selection while addressing challenges like collinearity, overfitting, and data imbalance to refine gradient-based metrics for analysis or heuristics.

\subsection{Numerical Findings}
\label{sec:Numerical_Findings}
To gain further insight into the client selection formulation in Sec.~\ref{sec:Logistic_Regression_Pairwise}, 
we consider the training of the last layer of VGG for $K=10$ clients, $T=15$ iterations, $J=\{1,2,5\}$ shards, and $R=10$ seeds. For each of the training settings, i.e., $(t,j,r)$, we consider the selection of all possible $K(K-1)$ pairs of users $k,k'$ and record (i) a set of pairwise gradient features $\mathbf{x}_{k,k'}(t,j,r)$, and (ii) the accuracy of the model under this selection $y'_{k,k'}(t,j,r)$, as discussed above. 

The features we consider are quite extensive, such as:
The features we consider include: 
(i) \emph{Per-user geometric features}, such as $\|\mathbf{g}\|_p + \|\mathbf{g}'\|_p$; 
(ii) \emph{Pairwise geometric features}, e.g., $\|\mathbf{g} - \mathbf{g}'\|_p$, $\cos_p(\mathbf{g}, \mathbf{g}')$; 
(iii) \emph{Per-user statistical features}, e.g., $\mathsf{Kurt}[\mathbf{g}] + \mathsf{Kurt}[\mathbf{g}']$; and 
(iv) \emph{Pairwise statistical features}, such as $\mathsf{Cov}(\mathbf{g}, \mathbf{g}')$. After careful experimentation, we conclude that the $\cos_p(\mathbf{g},\mathbf{g}')$ for $p \in [4]$ provides the most accurate and robust user selection performance. 

A summary of the training for this set of features is provided in Table~\ref{tab:log reg all feat}, where we report the performance in terms of (i) the rank of the features in the feature importance analysis and (ii) the relative loss, normalized over the loss of the worst user selection minus the loss of the user selection, across (i) shards and (ii) iterations. 
Note that the results in each row are averaged across the other training conditions. That is, the results for shard $j$ are averaged over the iteration $t$ and random seed $r$.
From the above, we glean the following two main insights:

\begin{itemize}[leftmargin=*]
    \item \textbf{Single-Feature Accuracy:} When restricting to a single feature for simplicity, PNCS consistently emerges as the most predictive and robust metric for successful client-pair selection. 
    While \(\cos_1\) achieves a smaller relative accuracy overall, it has a higher variance.
    
    \item \textbf{Negative Alignment:} Selecting the pair with the \emph{most negative} PNCS improves performance, suggesting that selecting complementary gradients—as opposed to overly similar ones—is beneficial in most settings. 
\end{itemize}
\subsection{Proposed Algorithm: PNCS}
\label{sec:cosl4-select}

\begin{figure*}[!h]
  \centering
 \includegraphics{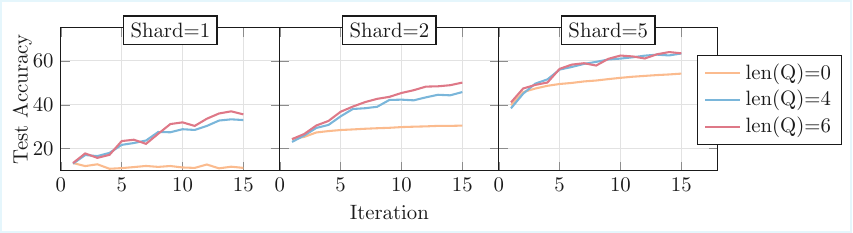}
  \caption{Comparison of PNCS with different queue lengths $L$ using dataset CIFAR-10 for shards 
  $j \in [1,2,5]$  under GSCCS settings.}
  \vspace{-0.25cm}
 \label{fig:ab_queue}
\end{figure*}

\begin{table}[ht]
    \centering
        \caption{Feature ranking and relative accuracy in logistic regression models  in Sec.~\ref{sec:Numerical_Findings}}
    \includegraphics{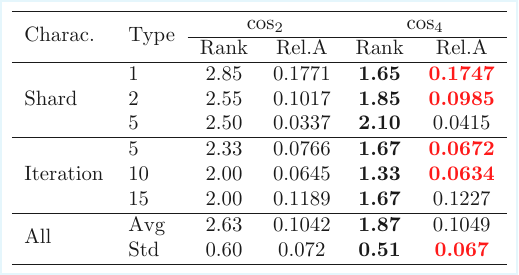}


    \label{tab:log reg all feat}
\end{table}
Having gleaned the insights from Section~\ref{sec:Numerical_Findings}, we now present our proposed solution for the GSCCS setting, which we refer to as PNCS. Before describing the full procedure, we introduce three additional elements that generalize and stabilize the selection process.

\noindent
\textbf{Generalizing Beyond Pairs.}
Although the earlier analysis focused on selecting exactly two clients, we naturally extend this to subsets of size $J > 2$. Rather than computing pairwise PNCS for two users, we consider the average similarity across all pairs in the subset $\mathcal{S}$:
\begin{equation}
\label{eq:cos_avg}
\overline{\cos_4}(\mathcal{S})
\;=\;
\frac{1}{\binom{J}{2}}
\sum_{\substack{k,k' \in \mathcal{S} \\ k > k'}}
\cos_4\bigl(\mathbf{g}_k,\mathbf{g}_{k'}\bigr).
\end{equation}
A lower value of \(\overline{\cos_4}(\mathcal{S})\) indicates a higher degree of gradient diversity, which can be beneficial in non-i.i.d.\ settings. We will demonstrate in subsequent sections that this multi-client criterion maintains the advantages of the pairwise approach while scaling to larger subsets.

\noindent
\textbf{Generalizing Beyond a Single Layer.}
Next, while Section~\ref{sec:Numerical_Findings} considered only one layer of the model, Equation~\eqref{eq:cos_avg} extends naturally to deeper networks. We simply compute each \(\cos_4(\mathbf{g}_k,\mathbf{g}_{k'})\) over multiple layers (or an appropriately weighted sum of per-layer similarities) and then aggregate those into an overall \(\overline{\cos_4}\) score. This allows our method to capture gradient diversity spanning the entire neural architecture, rather than focusing on a single layer.

\noindent
\textbf{AoU-Queue for Client Rotation.}
Finally, to avoid selecting the same clients repeatedly—especially in highly heterogeneous scenarios—we introduce an \emph{Age-of-Update Queue} (AoU-Queue).
Whenever a client is chosen for transmission, we place it in the AoU-Queue for a fixed number of rounds, preventing its immediate re-selection. Mathematically, let $L$ be the length of the queue. If a user is selected for transmission at time $t$, it will only be available for selection again at time $t'$ such that:
$t' > t + \frac{L}{S}$.


This introduces a "cool-down period" to sample less frequently chosen clients, forming the basis of the PNCS algorithm alongside multi-user selection, multi-layer gradients, and an AoU-Queue.
Each client $k$ first compresses its local gradient $\mathbf{g}_{kt}$ into a lower-dimensional sketch $\mathbf{s}_{kt} = \phi(\mathbf{g}_{kt})$ and sends it to the PS. Based on these summaries, the PS approximates the PNCS similarity for every pair of clients not currently in the queue. It then selects a subset $\mathcal{S}_t$ of size $J$ by \emph{maximizing negative alignment} (i.e., minimizing $\overline{PNCS}$). Those $J$ clients are enqueued for $\ell$ rounds to avoid repeated selection, while they transmit their \emph{full} gradients to the PS. Finally, the server aggregates these gradients, updates the global model, and broadcasts it back to all clients.

As shown in Fig. 1, using a nonzero queue length improves accuracy across shard configurations by balancing gradient-based selection with client rotation. Mixing a negative PNCS alignment criterion with a queue ensures diversity and fairness, leading to faster convergence and higher accuracy in heterogeneous FL settings.

\section{Experiments}
\label{sec:Experiments}

\begin{table*}[ht]
    \centering
    \caption{Comparison of PNCS with baselines across various settings, showing results for Layers 1/2.}
    \includegraphics{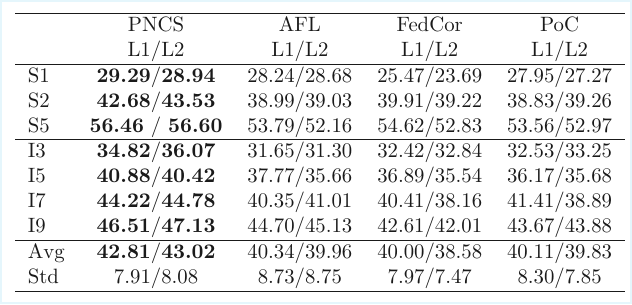}
\end{table*}

\subsection{Settings}
\label{sec:exp_setting}
Experiments are conducted on the CIFAR-10~\cite{cifar10} and Fashion-MNIST datasets~\cite{xiao2017_online} with an ImageNet pre-trained VGG16 network~\cite{vgg16}, where the feature extraction layers are frozen, and only the classifier layers are fine-tuned. Each experiment is repeated with \textbf{10} random seeds to ensure reliability.
We compare it against three baselines: FedCor~\cite{fedcor}, AFL~\cite{afl}, and POWER-OF-CHOICE~\cite{powd}.
The dataset is divided into $K \times S = 10 \times S$ shards, where $K=10$ represents the number of clients and $S$ is a hyperparameter controlling the level of heterogeneity (with lower $S$ indicating higher non-IIDness). Within each shard, data points share identical labels. Our method was evaluated under three heterogeneity levels ($S=1, 2, 5$) and two selection configurations: performing selection on only layer 6 (one-layer) and on layers 3 and 6 (two-layer) of the VGG classifier.


\subsection{Compare to SOTA}
\label{sec:comp_sota}
Test accuracy versus communication for the CIFAR10 experiments
\subsubsection{CIFAR-10 Dataset with 2 Clients Selected}
Table~\ref{tab:layer_comparison} demonstrates that our PNCS method consistently outperforms the baseline methods on \textbf{CIFAR-10} under different data heterogeneity ($\text{Shard}=1, 2, 5$).

At the top half of the table, we display the converged test accuracy of each method. The results show that PNCS converges to the best test accuracy compared to the baseline methods across all three heterogeneity settings. At the bottom half of the table, the test accuracy (averaged across three heterogeneity settings) at selected iterations is illustrated. Extending to the two-layer configuration, while the baseline methods generally result in a loss of performance, our PNCS method may achieve higher test accuracy.
\subsubsection{CIFAR-10 dataset with 4 clients selected}

\begin{figure}[ht]
  \centering
 \includegraphics{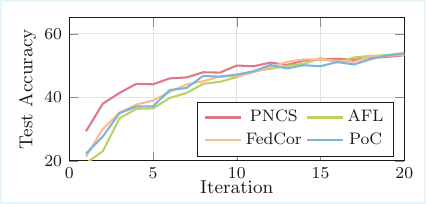}
  \caption{Comparison with baselines on the CIFAR-10 dataset for  shard number $J=2$, $S=4$, $K=10$,  in the one-layer setting.}
\vspace{-0.3cm}
 \label{fig:1layer_sel4_shard2_cifar}
\end{figure}

In Figure~\ref{fig:1layer_sel4_shard2_cifar}, we compare our method with the baseline methods when 4 out of 10 total clients need to be selected. 
Although all four methods converge to similar final test accuracies after 20 iterations, our method excels the other in terms of converging speed. 
While the baseline methods all take more than 5 iterations to reach 40\% test accuracy, our method achieves that at the third iteration. 
This advantage in convergence speed persists until our method is the first to achieve the final converged test accuracy.

\subsubsection{Fashion-MNIST dataset with 2 clients selected}

\begin{figure}[ht]
  \centering
 \includegraphics{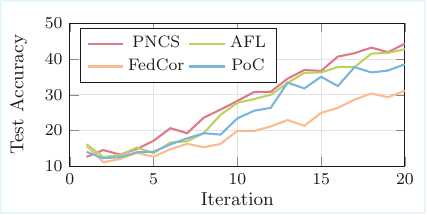}
  \caption{Comparison with baselines on the Fashion-MNIST dataset ($J = 1,S = 2, K=10$ ) with one-layer setting.}

 \label{fig:fasion_mnist}
\end{figure}

To validate the effectiveness of our method on other datasets, we deploy all three baselines and PNCS on \textbf{Fashion-MNIST} dataset.
In Figure~\ref{fig:fasion_mnist}, we present the experiment results under the setting, where local dataset is highly heterogeneous (Shard =1) and 2 out of 10 clients need to be selected for the one-layer setting.
By analyzing the training curves, we observe that our PNCS method starts to outperform the baseline methods starting from the fourth iteration, and our method converges to the best final test accuracy among all comparisons.


\subsection{Ablation Study on Queue}
\label{sec:ab_queue}

Figure ~\ref{fig:ab_queue} shows that a queue length of Q=4 optimally balances stability and adaptability, achieving faster convergence in heterogeneous data (small shards) and slightly improving early convergence in homogeneous data (large shards), while longer queues (Q=6) underperform due to diminishing returns.


\bibliographystyle{IEEEtran}
\bibliography{ref}

\begin{thebibliography}{10}
\providecommand{\url}[1]{#1}
\csname url@samestyle\endcsname
\providecommand{\newblock}{\relax}
\providecommand{\bibinfo}[2]{#2}
\providecommand{\BIBentrySTDinterwordspacing}{\spaceskip=0pt\relax}
\providecommand{\BIBentryALTinterwordstretchfactor}{4}
\providecommand{\BIBentryALTinterwordspacing}{\spaceskip=\fontdimen2\font plus
\BIBentryALTinterwordstretchfactor\fontdimen3\font minus \fontdimen4\font\relax}
\providecommand{\BIBforeignlanguage}[2]{{%
\expandafter\ifx\csname l@#1\endcsname\relax
\typeout{** WARNING: IEEEtran.bst: No hyphenation pattern has been}%
\typeout{** loaded for the language `#1'. Using the pattern for}%
\typeout{** the default language instead.}%
\else
\language=\csname l@#1\endcsname
\fi
#2}}
\providecommand{\BIBdecl}{\relax}
\BIBdecl

\bibitem{mcmahan2017federated}
B.~McMahan and D.~Ramage, ``Federated learning: Collaborative machine learning without centralized training data,'' \emph{Google Research Blog}, vol.~3, 2017.

\bibitem{li2020convergencefedavgnoniiddata}
\BIBentryALTinterwordspacing
X.~Li, K.~Huang, W.~Yang, S.~Wang, and Z.~Zhang, ``On the convergence of fedavg on non-iid data,'' 2020. [Online]. Available: \url{https://arxiv.org/abs/1907.02189}
\BIBentrySTDinterwordspacing

\bibitem{konevcny2016federated}
J.~Kone{\v{c}}n{\`y}, ``Federated learning: Strategies for improving communication efficiency,'' \emph{arXiv preprint arXiv:1610.05492}, 2016.

\bibitem{mcmahan2017communication}
B.~McMahan, E.~Moore, D.~Ramage, S.~Hampson, and B.~A. y~Arcas, ``Communication-efficient learning of deep networks from decentralized data,'' in \emph{Artificial intelligence and statistics}.\hskip 1em plus 0.5em minus 0.4em\relax PMLR, 2017, pp. 1273--1282.

\bibitem{ma2022state}
X.~Ma, J.~Zhu, Z.~Lin, S.~Chen, and Y.~Qin, ``A state-of-the-art survey on solving non-iid data in federated learning,'' \emph{Future Generation Computer Systems}, vol. 135, pp. 244--258, 2022.

\bibitem{li2024comprehensive}
J.~Li, T.~Chen, and S.~Teng, ``A comprehensive survey on client selection strategies in federated learning,'' \emph{Computer Networks}, p. 110663, 2024.

\bibitem{lai2021oort}
F.~Lai, X.~Zhu, H.~V. Madhyastha, and M.~Chowdhury, ``Oort: Efficient federated learning via guided participant selection,'' in \emph{15th $\{$USENIX$\}$ Symposium on Operating Systems Design and Implementation ($\{$OSDI$\}$ 21)}, 2021, pp. 19--35.

\bibitem{sultana2022eiffel}
A.~Sultana, M.~M. Haque, L.~Chen, F.~Xu, and X.~Yuan, ``Eiffel: Efficient and fair scheduling in adaptive federated learning,'' \emph{IEEE Transactions on Parallel and Distributed Systems}, vol.~33, no.~12, pp. 4282--4294, 2022.

\bibitem{balakrishnan2022diverse}
R.~Balakrishnan, T.~Li, T.~Zhou, N.~Himayat, V.~Smith, and J.~Bilmes, ``Diverse client selection for federated learning via submodular maximization,'' in \emph{International Conference on Learning Representations}, 2022.

\bibitem{nguyen2020fast}
H.~T. Nguyen, V.~Sehwag, S.~Hosseinalipour, C.~G. Brinton, M.~Chiang, and H.~V. Poor, ``Fast-convergent federated learning,'' \emph{IEEE Journal on Selected Areas in Communications}, vol.~39, no.~1, pp. 201--218, 2020.

\bibitem{9283151}
L.~Huang, L.~Feng, H.~Wang, Y.~Hou, K.~Liu, and C.~Chen, ``A preliminary study of improving evolutionary multi-objective optimization via knowledge transfer from single-objective problems,'' in \emph{2020 IEEE International Conference on Systems, Man, and Cybernetics (SMC)}, 2020, pp. 1552--1559.

\bibitem{10219820}
Y.~Shi, Z.~Liu, Z.~Shi, and H.~Yu, ``Fairness-aware client selection for federated learning,'' in \emph{2023 IEEE International Conference on Multimedia and Expo (ICME)}, 2023, pp. 324--329.

\bibitem{battiloro2022lyapunov}
C.~Battiloro, P.~Di~Lorenzo, M.~Merluzzi, and S.~Barbarossa, ``Lyapunov-based optimization of edge resources for energy-efficient adaptive federated learning,'' \emph{IEEE Transactions on Green Communications and Networking}, vol.~7, no.~1, pp. 265--280, 2022.

\bibitem{zhou2023joint}
Z.~Zhou, S.~Shi, F.~Wang, Y.~Zhang, and Y.~Li, ``Joint client selection and cpu frequency control in wireless federated learning networks with power constraints,'' \emph{Entropy}, vol.~25, no.~8, p. 1183, 2023.

\bibitem{mehta2023greedy}
M.~Mehta and C.~Shao, ``A greedy agglomerative framework for clustered federated learning,'' \emph{IEEE Transactions on Industrial Informatics}, vol.~19, no.~12, pp. 11\,856--11\,867, 2023.

\bibitem{9317862}
S.~Zhai, X.~Jin, L.~Wei, H.~Luo, and M.~Cao, ``Dynamic federated learning for gmec with time-varying wireless link,'' \emph{IEEE Access}, vol.~9, pp. 10\,400--10\,412, 2021.

\bibitem{liu2023mflces}
Z.~Liu, S.~Duan, S.~Wang, Y.~Liu, and X.~Li, ``Mflces: Multi-level federated edge learning algorithm based on client and edge server selection,'' \emph{Electronics}, vol.~12, no.~12, p. 2689, 2023.

\bibitem{mohammed2020budgeted}
I.~Mohammed, S.~Tabatabai, A.~Al-Fuqaha, F.~El~Bouanani, J.~Qadir, B.~Qolomany, and M.~Guizani, ``Budgeted online selection of candidate iot clients to participate in federated learning,'' \emph{IEEE Internet of Things Journal}, vol.~8, no.~7, pp. 5938--5952, 2020.

\bibitem{chen2019pre}
M.~Chen, K.~Ho, Y.~Hung, C.~Su, C.~Kuan, H.~Tai, N.~Cheng, and C.~Lin, ``Pre-treatment quality of life as a predictor of distant metastasis-free survival and overall survival in patients with head and neck cancer who underwent free flap reconstruction,'' \emph{European Journal of Oncology Nursing}, vol.~41, pp. 1--6, 2019.

\bibitem{9372789}
P.~Zhang, C.~Wang, C.~Jiang, and Z.~Han, ``Deep reinforcement learning assisted federated learning algorithm for data management of iiot,'' \emph{IEEE Transactions on Industrial Informatics}, vol.~17, no.~12, pp. 8475--8484, 2021.

\bibitem{albelaihi2023deep}
R.~Albelaihi, A.~Alasandagutti, L.~Yu, J.~Yao, and X.~Sun, ``Deep-reinforcement-learning-assisted client selection in nonorthogonal-multiple-access-based federated learning,'' \emph{IEEE Internet of Things Journal}, vol.~10, no.~17, pp. 15\,515--15\,525, 2023.

\bibitem{arulkumaran2017deep}
K.~Arulkumaran, M.~P. Deisenroth, M.~Brundage, and A.~A. Bharath, ``Deep reinforcement learning: A brief survey,'' \emph{IEEE Signal Processing Magazine}, vol.~34, no.~6, pp. 26--38, 2017.

\bibitem{9991371}
C.~Wang, Q.~Wu, Q.~Ma, and X.~Chen, ``A buffered semi-asynchronous mechanism with mab for efficient federated learning,'' in \emph{2022 International Conference on High Performance Big Data and Intelligent Systems (HDIS)}, 2022, pp. 180--184.

\bibitem{zhu2022online}
H.~Zhu, Y.~Zhou, H.~Qian, Y.~Shi, X.~Chen, and Y.~Yang, ``Online client selection for asynchronous federated learning with fairness consideration,'' \emph{IEEE Transactions on Wireless Communications}, vol.~22, no.~4, pp. 2493--2506, 2022.

\bibitem{wang2020novel}
Y.~Wang and B.~Kantarci, ``A novel reputation-aware client selection scheme for federated learning within mobile environments,'' in \emph{2020 IEEE 25th International Workshop on Computer Aided Modeling and Design of Communication Links and Networks (CAMAD)}.\hskip 1em plus 0.5em minus 0.4em\relax IEEE, 2020, pp. 1--6.

\bibitem{batool2023block}
Z.~Batool, K.~Zhang, and M.~Toews, ``Block-racs: Towards reputation-aware client selection and monetization mechanism for federated learning,'' \emph{ACM SIGAPP Applied Computing Review}, vol.~23, no.~3, pp. 49--65, 2023.

\bibitem{wang2021collective}
Y.~Wang, Z.~Cao, D.~D. Zeng, Q.~Zhang, and T.~Luo, ``The collective wisdom in the covid-19 research: Comparison and synthesis of epidemiological parameter estimates in preprints and peer-reviewed articles,'' \emph{International Journal of Infectious Diseases}, vol. 104, pp. 1--6, 2021.

\bibitem{tan2022reputation}
X.~Tan, W.~C. Ng, W.~Y.~B. Lim, Z.~Xiong, D.~Niyato, and H.~Yu, ``Reputation-aware federated learning client selection based on stochastic integer programming,'' \emph{IEEE Transactions on Big Data}, 2022.

\bibitem{afl}
\BIBentryALTinterwordspacing
J.~Goetz, K.~Malik, D.~Bui, S.~Moon, H.~Liu, and A.~Kumar, ``Active federated learning,'' 2019. [Online]. Available: \url{https://arxiv.org/abs/1909.12641}
\BIBentrySTDinterwordspacing

\bibitem{powd}
\BIBentryALTinterwordspacing
Y.~Jee~Cho, J.~Wang, and G.~Joshi, ``Towards understanding biased client selection in federated learning,'' in \emph{Proceedings of The 25th International Conference on Artificial Intelligence and Statistics}, ser. Proceedings of Machine Learning Research, G.~Camps-Valls, F.~J.~R. Ruiz, and I.~Valera, Eds., vol. 151.\hskip 1em plus 0.5em minus 0.4em\relax PMLR, 28--30 Mar 2022, pp. 10\,351--10\,375. [Online]. Available: \url{https://proceedings.mlr.press/v151/jee-cho22a.html}
\BIBentrySTDinterwordspacing

\bibitem{marnissi2024client}
O.~Marnissi, H.~E. Hammouti, and E.~H. Bergou, ``Client selection in federated learning based on gradients importance,'' in \emph{AIP Conference Proceedings}, vol. 3034, no.~1.\hskip 1em plus 0.5em minus 0.4em\relax AIP Publishing, 2024.

\bibitem{sivasubramanian2024gradient}
D.~Sivasubramanian, L.~Nagalapatti, R.~Iyer, and G.~Ramakrishnan, ``Gradient coreset for federated learning,'' in \emph{Proceedings of the IEEE/CVF Winter Conference on Applications of Computer Vision}, 2024, pp. 2648--2657.

\bibitem{xu2023federated}
Y.~Xu, Z.~Jiang, H.~Xu, Z.~Wang, C.~Qian, and C.~Qiao, ``Federated learning with client selection and gradient compression in heterogeneous edge systems,'' \emph{IEEE Transactions on Mobile Computing}, 2023.

\bibitem{fedcor}
M.~Tang, X.~Ning, Y.~Wang, J.~Sun, Y.~Wang, H.~Li, and Y.~Chen, ``Fedcor: Correlation-based active client selection strategy for heterogeneous federated learning,'' in \emph{Proceedings of the IEEE/CVF Conference on Computer Vision and Pattern Recognition}, 2022, pp. 10\,102--10\,111.

\bibitem{cifar10}
\BIBentryALTinterwordspacing
A.~Krizhevsky, V.~Nair, and G.~Hinton, ``Cifar-10 (canadian institute for advanced research),'' 2009. [Online]. Available: \url{http://www.cs.toronto.edu/~kriz/cifar.html}
\BIBentrySTDinterwordspacing

\bibitem{xiao2017_online}
H.~Xiao, K.~Rasul, and R.~Vollgraf, ``Fashion-{MNIST}: a novel image dataset for benchmarking machine learning algorithms,'' 2017.

\bibitem{vgg16}
\BIBentryALTinterwordspacing
K.~Simonyan and A.~Zisserman, ``Very deep convolutional networks for large-scale image recognition,'' 2015. [Online]. Available: \url{https://arxiv.org/abs/1409.1556}
\BIBentrySTDinterwordspacing

\end{thebibliography}

\end{document}